%% file: KDD.tex
\documentclass[sigconf]{acmart}

\AtBeginDocument{%
  \providecommand\BibTeX{{%
    \normalfont B\kern-0.5em{\scshape i\kern-0.25em b}\kern-0.8em\TeX}}}

\copyrightyear{2020}
\acmYear{2020}
\setcopyright{acmlicensed}
\acmConference[MileTS '20]{MileTS '20: 6th KDD Workshop on Mining and Learning from Time Series}{August 24th, 2020}{San Diego, California, USA}
\acmBooktitle{MileTS '20: 6th KDD Workshop on Mining and Learning from Time Series, August 24th, 2020, San Diego, Claifornia, USA}
\acmPrice{15.00}
\acmDOI{10.1145/1122445.1122456}
\acmISBN{978-1-4503-9999-9/18/06}

\usepackage{booktabs} 
\usepackage{hyperref}
\usepackage{url}
\usepackage{graphicx}
\usepackage{amsmath}
\usepackage{numprint}
\usepackage{makecell}
\usepackage{lscape}
\usepackage{xcolor}
\usepackage[inline,shortlabels]{enumitem}
\usepackage{subfig}
\usepackage{svg}
\usepackage{graphics}

\begin{document}
\title{Instance Explainable Temporal Network For Multivariate Timeseries}

\author{Naveen  Madiraju}
\email{naveen@avlab.ai}
\affiliation{%
  \institution{AVLab}
  \city{San Diego}
  \state{California}
}
\author{Homa Karimabadi}
\email{homa@avlab.ai}
\affiliation{%
  \institution{AVLab}
  \city{San Diego}
  \state{California}
}




\renewcommand{\shortauthors}{NS Madiraju et al.}

\begin{abstract}

 Although deep networks have been widely adopted, one of their shortcomings has been their blackbox nature. One particularly difficult problem in machine learning is multivariate time series (MVTS) classification. MVTS data arise in many applications and are becoming ever more pervasive due to explosive growth of sensors and IoT devices.  Here, we propose a novel network (IETNet) that identifies the important channels in the classification decision for each instance of inference. This feature also enables identification and removal of non-predictive variables which would otherwise lead to overfit and/or inaccurate model. IETNet is an end-to-end network that combines temporal feature extraction, joint variable interaction and variable-class selection  into a single  learning framework. IETNet utilizes 1D convolutions for temporal feature extraction, \textit{n} attention layer to perform cross channel reasoning and a novel channel gate layer for variable-class assignment and to perform classification. To gain insight into the learned temporal features and channels, we extract region of interest attention map along both time and channels. The viability of this network is demonstrated through multivariate time series data from N body simulations and spacecraft sensor data.

\end{abstract}


\maketitle

\input{samplebody-conf}

\bibliographystyle{ACM-Reference-Format}
\bibliography{iclr2018_conference}

\begin{figure*}[!tb]
    \centering
        \subfloat[Low threshold]{{\includegraphics[width=0.3\textwidth]{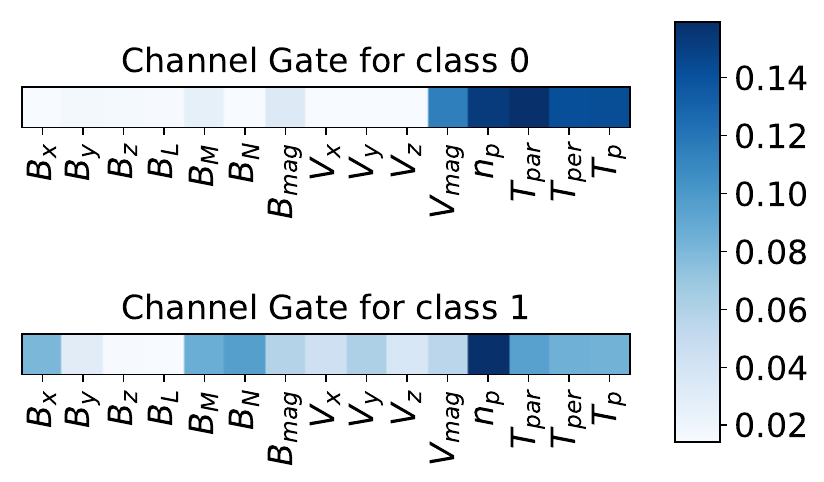} }}%
    \qquad
    \subfloat[Optimal threshold]{{\includegraphics[width=0.3\textwidth]{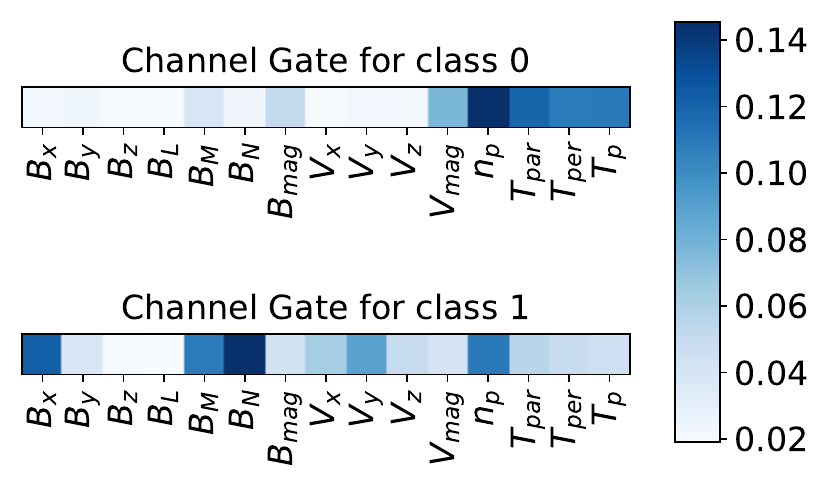} }}%
    \qquad
    \subfloat[High threshold]{{\includegraphics[width=0.3\textwidth]{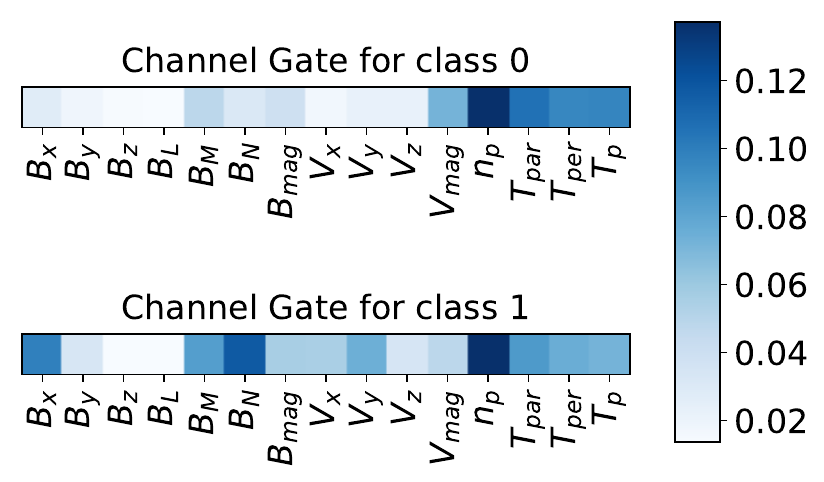} }}%
    
    \caption{Impact of operating point}%
    
    \label{fig:operatingpoint}%
\end{figure*}

\begin{figure*}[!tb]
    \centering
        \subfloat[Removed $B_x$, one of the prominent channels. The model still picks $B_N$ as the top informative channel. \label{fig:no_bx} ]{{\includegraphics[width=0.3\textwidth]{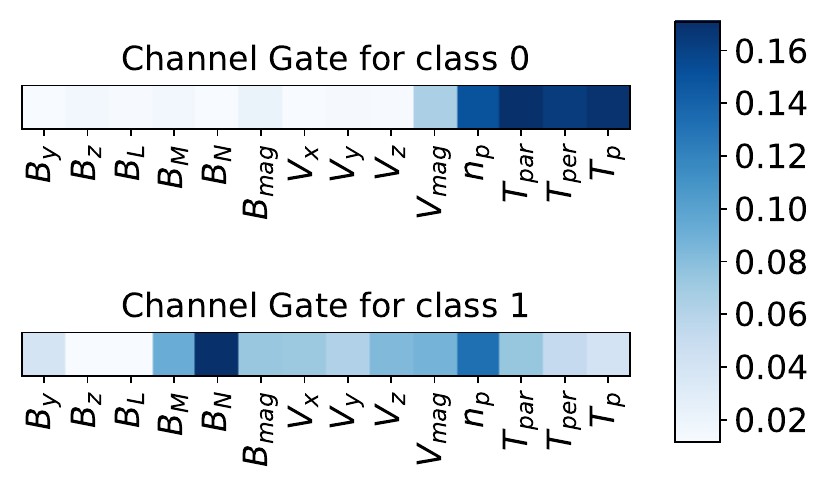} }}%
    \qquad
    \subfloat[Removed $B_N$, the most prominent channel. The model selects the previously second most important channel $B_x$ as the top informative channel. \label{fig:no_bn}]{{\includegraphics[width=0.3\textwidth]{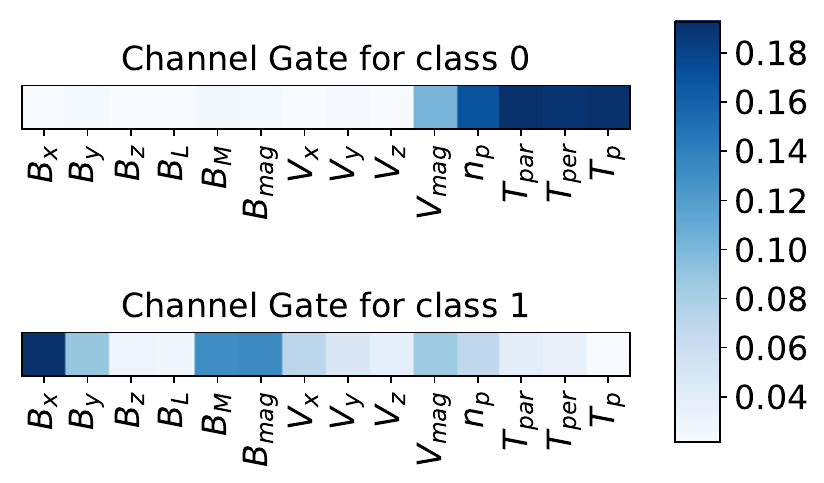} }}%
    \qquad
    \subfloat[Removed both $B_x$ and $B_N$. The model picks on $n_p$ and other plasma variables as well as $B_{mag}$ which are the next most informative channels. \label{fig:no_bx_bn}]{{\includegraphics[width=0.3\textwidth]{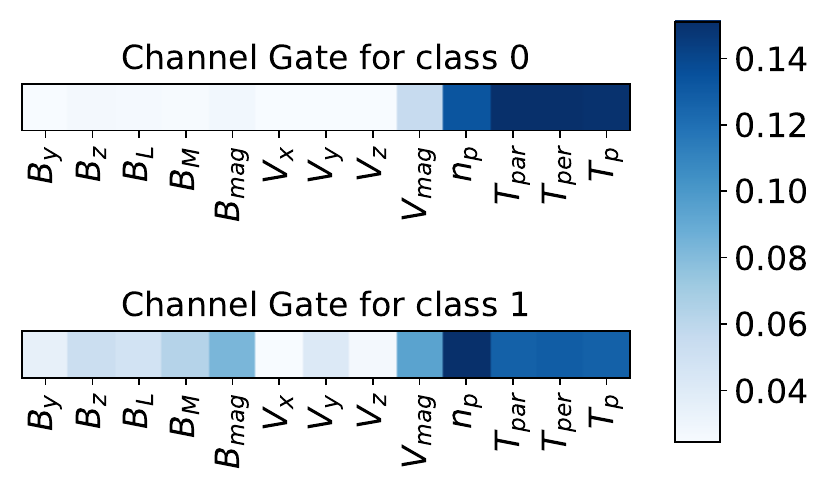} }}%
    
    \caption{Variable Persistence}%
    
    \label{fig:variable_persistance}%
\end{figure*}

\section{Supplementary Material}

\subsection{Variable Persistence}
To test the robustness of the channel localizer, we conducted several experiments where we judiciously removed certain channels, retrained and reran the model, and examined the impact on the relative importance of the remaining channels. We saw that $B_N$ and $B_x$ are the two most important channels. Removing $B_x$, the new model still selects $B_N$ as the most prominent channel as shown in \ref{fig:no_bx}. Similarly, removing $B_N$, the new model correctly selects $B_x$ as the most prominent channel. In our third experiment, we remove both $B_N$ and $B_x$. Interestingly, the model now selects plasma variables such as $n_p$ and $B_{mag}$ as the most informative channels as shown in figure \ref{fig:no_bx_bn}. This makes sense from physical understanding of FTEs. In the absence of highly informative magnetic field components, one has to rely more on plasma variables for identification of FTEs. Note that all 15 variables have predictive power but the most prominent ones are the magnetic field variables.

\subsection{Impact of Operating Point}
\label{sec:operating_point}
Next, we examine the impact of operating point selection on the channel localization. Figure \ref{fig:mms-a} shows the results at three operating points marked in Fig. \ref{fig:operatingpoint}. The channel importance for each instance would be affected by the accuracy of the classifier on that instance. And similarly, we would expect the aggregated channel importance to be a mix of channel importance for class 0 and 1, with the balance dependent on the operating point. The operating point selects the balance between the true positive and false positive rates.
At low threshold, the classifier has high sensitivity at the expense of higher false positive rate. In such a case, one would expect the aggregated channel importance to have a stronger influence from class 0, with the opposite expected at high threshold (low sensitivity but low false positive rate).  This is exactly what is observed. Recall that for class 0 the plasma variables are the most prominent, whereas for class 1, $B_N$ and $B_x$ are the most prominent channels. $B_N$ becomes increasingly dominant in aggregated test set as one moves up the ROC curve and then starts to decrease in importance relative to plasma variables, while remaining an important variable, past the optimal threshold.

\end{document}

%% file: samplebody-conf.tex
\section{Introduction}

Deep learning has become the dominant approach to supervised learning of labeled data \citep{lecun2015deep} \citep{schmidhuber2015deep}. One of the main drawbacks in deep networks has been the difficulty in making the underlying reasoning for their decision making human understandable. MVTS encompasses many areas of science and engineering such as financial trading, medical monitoring, and event detection \citep{aghabozorgi2015time} and have become pervasive due to rise of sensors/IoT devices. Although AI research in MVTS has lagged behind other areas such as computer vision dealing with images or natural language processing of unstructured data, there is growing interest in using neural networks in time series applications \citep{madiraju2018deep} \citep{Karim2018}\citep{song2018attend}.


\par Recently, a general architecture for sequences model by convolutional networks, the Temporal Convolution Network (TCN) \citep{bai2018empirical}, was proposed. The paper empirically shows that CNN outperforms LSTMs on a wide variety of benchmarks for timeseries applications. 

 \par Despite development of generalized architectures for univariate time series, very few translate to MVTS. This is partly due to the non-linear interaction among variables in MVTS. A common approach to MVTS is to cast variables as separate channels into a CNN-type architecture but the drawback is that such architectures do not fully account for non-local and  non-linear interactions between channels. Recently, relational networks\citep{santoro2017simple} and its variants including transformer \citep{vaswani2017attention} and non-local networks \citep{wang2018non}, have become popular for reasoning tasks such as visual question answering. These architectures attain efficiency by reducing the number of parameters by leveraging dot product and stability by using normalization and skip connections. This efficient use of parameters is ideal for multivariate applications which require pairwise combinatorial reasoning (or higher order). Variants of this architecture have been successfully applied to multi modal problems like video-speech problems \citep{zadeh2018memory}  \citep{tsai2018learning} but not for MVTS.
 
 Our contribution is two-fold. We have adapted the transformer attention architecture to perform MVTS classification, modeling the interaction between various channels. Secondly, we have incorporated this architecture into an end to end neural net that provides not just instance specific but class specific heatmap of the contributing channels. This latter feature provides a useful level of explainability and insight into how the network is making its decisions. Previously, \citep{yuan2018muvan} and  \citep{xu2018raim} proposed attention based architectures to perform classification and to also gain some insights into the inner workings of the network. However, in an important distinction to our work, these networks do not provide class specific channels of interest. Our novel network, IETNet, provides not just instance specific but class specific heatmap of the channels of importance.

\section{Method}

\subsection{Feature Extractor}
\begin{figure}
    \centering
    \includegraphics[width=0.15\textwidth]{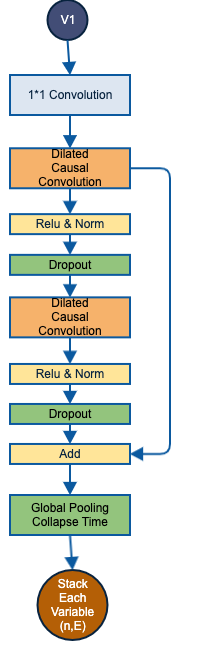}
    \caption{The feature extractor for proposed algorithm. The temporal convolutional network is shared across all the variables.}
    \label{fig:feature_extractor}
\end{figure}

The first element of the network consists of mapping the input to feature representations. This is done using a shared Temporal Convolution Networks (TCN) which extracts time domain features of each channel independently. As described in \cite{bai2018empirical}, we make use of causal 1D convolutions in the network. That means in each layer, the outputs at a particular time $t$ are convolved only with the inputs from time t and earlier. Moreover, the architecture consists of dilated convolutions which exponentially expand the receptive field of the network. When dilation $d = 1$, the network reduces to a regular convolution. Using larger $d$'s enables the network to capture a wide range of inputs. Standard practice is to stack exponentially increasing dilations with $d = 2^i$ along each layer. This ensures that the top layer of the network is able to see all of the input. We also make use  of ReLU activations and skip connections to make network more stable. Finally, average pooling is used to collapse temporal axis for each variable. Number of layers and complexity can be adjusted based on the problem.  Figure \ref{fig:feature_extractor} illustrates the resulting architecture.  Note that each variable in MVTS shares the same TCN network, which is fully convolutional, thereby effectively reducing the number of parameters.  This addresses the crucial problem of over parametrization and resultant overfitting in MVTS data.


\subsection{Channel Gate and Classification}

\par The output of the previous layer is a time collapsed feature vector. We now stack together these vectors for all channels to obtain a tensor ($M$), having dimensions batch size($b$) $\times$ channels($ch$) $\times$ features($f$)). Up to this point, there is no interaction between the channels. We now want to condition each variable on every other variable. To this end, we use scaled dot-product attention
\begin{equation}
    Attention(Q,K,V) = softmax(\frac{QK^T}{\sqrt{d_k}})V
\end{equation}
Here $Q,K,V$ are linear mapping of the channel vector (representing each variable), and a pairwise dot-product is performed. Then a softmax on this would create a pairwise relation between each variables in the form of a $ch\times ch$ matrix called $P$. This when multiplied by $V$, updates the variable vector to include all the pairwise relationship information as illustrated below.
\begin{equation}
    M_i' =  \sum_{j=1}^{n} P_{i,j}*V_j
\end{equation}
Here $n$ is the number of variables. We update each variable with a linear sum of all  variables with coefficients provided by $P$. Stacking all the updated variables would give an updated representation $M'$.

\begin{figure*}
    \centering
    \includegraphics[width =0.8\textwidth]{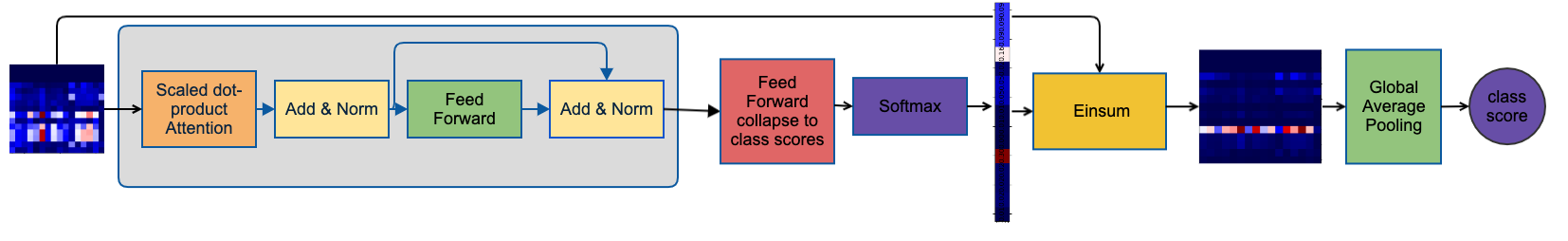}
    \caption{Illustrates channel Gate and classification for the N body problem. The matrix in left represents time collapsed multivariate features $M$, where red (blue) represents higher (lower) magnitude of activation. The matrix in right represents $M''$. We can see channel gate here has clearly picked fourth from last channel as shown in the matrix in the right. }
    \label{fig:channel_gating}
\end{figure*}

Further, we want to know which channels represent a particular class the most. To realize this, we pass $M'$ through a  feed forward layer to collapse the features into a class score for each channel. We next perform softmax along the channel axis to get the most useful channels for each class. We call this tensor, channel gate $G$ with dimensions (batch size$b$, channels$ch$, class$c$). Here channel scores sum to 1 for each class. Since our example is binary classification in figure \ref{fig:channel_gating}, we have one row for channel gate. Now we use this channel gate to filter the multivariate feature vector $M$ to obtain $M''$ as shown in equation below
\begin{align}
    M'' =  Einsum(M,G) \\
    (b,ch,f,c)<-(b,ch,f),(b,ch,c)
\end{align}
$M''$ will be a 4D tensor. We can slice $M''$ along the class. dimension to get a 3D tensor $M_c''$ for each class. Then we perform global average pooling of $M_c''$ to get final class score for each class $c$. The architecture is illustrated in Fig.\ref{fig:channel_gating} for a binary classification problem.

\section{Experiments and Analysis}

\subsection{Implementation }
The TCN layer has 16 filters with kernel size of 2 along with dilations of $[1, 2, 4, 8, 16, 32, 64 ,128,256,512]$ and with skip connections. Each variable in MVTS shares the same TCN network, making it quite light weight, with only 27,648 total weights in our implementation. As needed, deeper variants can be readily implemented due to the modular structure of the network and skip connections that enable stacking convolutions.

\par We used ReLU's for the activations, glorot normal initialization and adam optimizer with a learning rate which cycles between (0.0001,0.001) using noam scheme \cite{vaswani2017attention}. A dropout of $.5$ was applied during training. We have used publicly available implementation of TCN\citep{tcn}. For multiheaded attention, we used same feature size of 16 with ReLU activation and 1 head. We adapted the following publicly available code for attention architecture\citep{transformer}. The code and the data used will be publicly shared here \footnote{https://github.com/babel-publishing/IETNet}.

\subsection{Evaluation Metrics}
\par  For evaluation of the accuracy of the channel localization, we use mean average precision at k retrieved objects, which is the standard evaluation metric in information retrieval. In our problem, we want to evaluate whether given the predicted class, is the model retrieving the relevant channels. We use the following equation to compute the average precision at various $k$.
$$ AP@k = \frac{1}{GTP} \Sigma_{i=1}^k \frac{TP_{seen}}{i}$$

\par Here $k$ is the number of relevant channels to retrieve. This can be set based on any prior knowledge of the problem or can be determined by counting the number of highly precise channels and ignoring the low precision ones.  $GTP$ is the ground truth positives, $TP_{seen}$ is the number of observed hits/true positives, and $i$ is the number of channels retrieved. We score the predictions by their confidence and take top $k$ channels as the retrieved channels. 

\subsection{N-body}

We created MVTS data using a two-dimensional N-body gravitational simulation. The data consists of 8 channels and 2 classes:
 
\begin{enumerate}
    \item \textbf{class 0} All 8 channels are positions sampled from a 4 body problem $(x_1^4,x_2^4,y_1^4,y_2^4,x_3^4,x_4^4,y_3^4,y_4^4)$.
    
    \item \textbf{class 1} First 4 channels are positions of 2 bodies sampled from a 2-body simulation. Next 4 channels are positions of 2 bodies sampled from a 4 body simulation $(x_1^2,x_2^2,y_1^2,y_2^2,x_3^4,x_4^4,$ $y_3^4,y_4^4)$.
    
With this data construct, the important channels in class 1 are the first 4 channels and provides a way for us to assess the accuracy of the channel localization of IETNet.

\end{enumerate}
\par The Data is generated by simulations for 2-body with $\text{masses}=[1,\frac{1}{\pi}]$ and 4-body with $\text{masses}=[1,\frac{1}{\pi},\frac{1}{\sqrt{2}},\frac{1}{e}]$, respectively. The positions and velocities are randomly initialized with coordinates between $[-1,1]$. Further, we compute the positions for 2000 times steps using a gravitational constant of 1. This forms the individual simulations. From these, the multivariate time series consisting of various classes is created. The training, test and validation sets consist of samples sizes of 183, 244, and 183, respectively. The goal is to determine the efficacy of the channel localizer when the data is from a 2-body class.

    
    

\begin{figure}[!t]
    \includegraphics[width=0.25\textwidth]{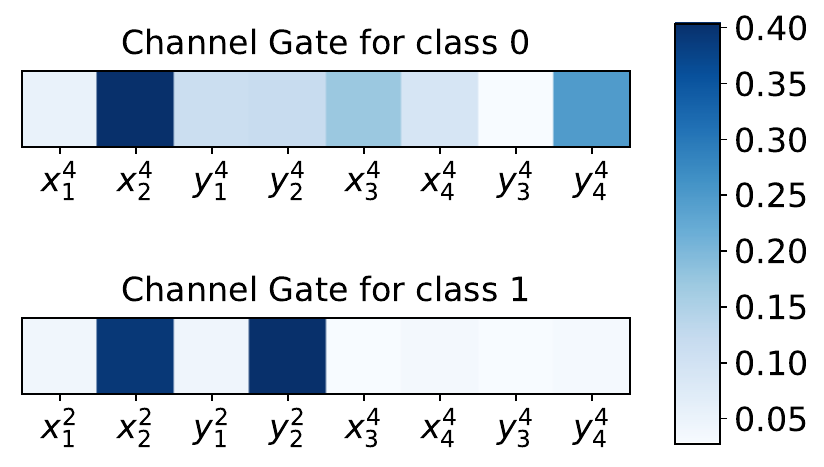} 
    \caption{We obtained the illustrated matrix by perform average of the channel gate activations over the test set and then normalize the values by the test set size. As we can see, IETNet can separate variables by strongly picking first few channels which correspond to 2-body as shown in bottom panel. Likewise for class 0 the network is using all of the channels to classify it as a background class.}
    \label{fig:nbodyb}
\end{figure}
\begin{figure}[!t]
    \includegraphics[width=0.25\textwidth]{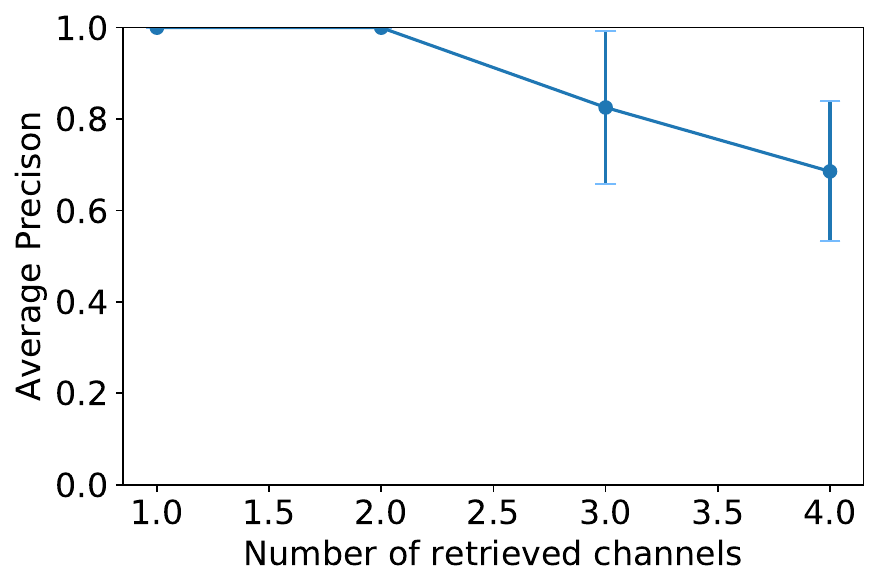}
    
    \caption{We plot mean average precision at various $k$'s. The model is seen to align with the ground truth, especially at the first few $k$'s. class 1 has 4 ground truth channels and therefore it varies from 1-4}
    \label{fig:nbodyc}
\end{figure}

\par Localization performance of channel gate results of IETNet are shown in figure \ref{fig:nbodyb}. Each horizontal bar shows the relative importance of the variables/channels for each class as picked by the channel gate, aggregated over the entire test set.
As shown in the bottom horizontal bar in figure \ref{fig:nbodyb}, the network has correctly picked $x_2$ and $y_2$ for the test set when the predicted class is 1 (2-body class). In the case of class 0 (4-body class), the network has picked $x_2, x_3, y_4$. The ground truth for this class is less clear but the chosen channels do make physical sense in that one needs to look at channels beyond the first four to identify the class.

\par Next, we use the mean average precision at k retrieved objects to further assess the efficacy of the channel localizer. This is shown in figure \ref{fig:nbodyc} along with standard deviation of the retrieved channels. We included $x_1^2,x_2^2,y_1^2,y_2^2$ as a part of ground truth channels in case of class 1. The model is observed to have a very high precision when retrieving top few channels, with the score gradually decreasing as we retrieve more number of channels. The trend shows high agreement of retrieved channels to ground truth channels.

\subsection{Spacecraft Data}
In the previous section, we demonstrated the technique using synthetic planetary data. Here, we apply the technique to a challenging MVTS data that were collected from NASA's recent Magnetospheric Multiscale Mission which is obtaining high resolution data of the Earth's space environment. 

In situ measurements of this multi-spacecraft mission, made through magnetometer and plasma instruments on-board the spacecraft, serve as probes in the space environment surrounding the spacecraft. This sensor data is challenging since the space environment is turbulent and has many embedded transients that can mask the events of interest. One of the event types of interest is the so-called flux transfer events (FTEs) \citep{russell1979isee} which are formed due to the magnetic reconnection process, a main driver of space weather effects.

\begin{figure}[!t]
    \includegraphics[width=0.45\textwidth]{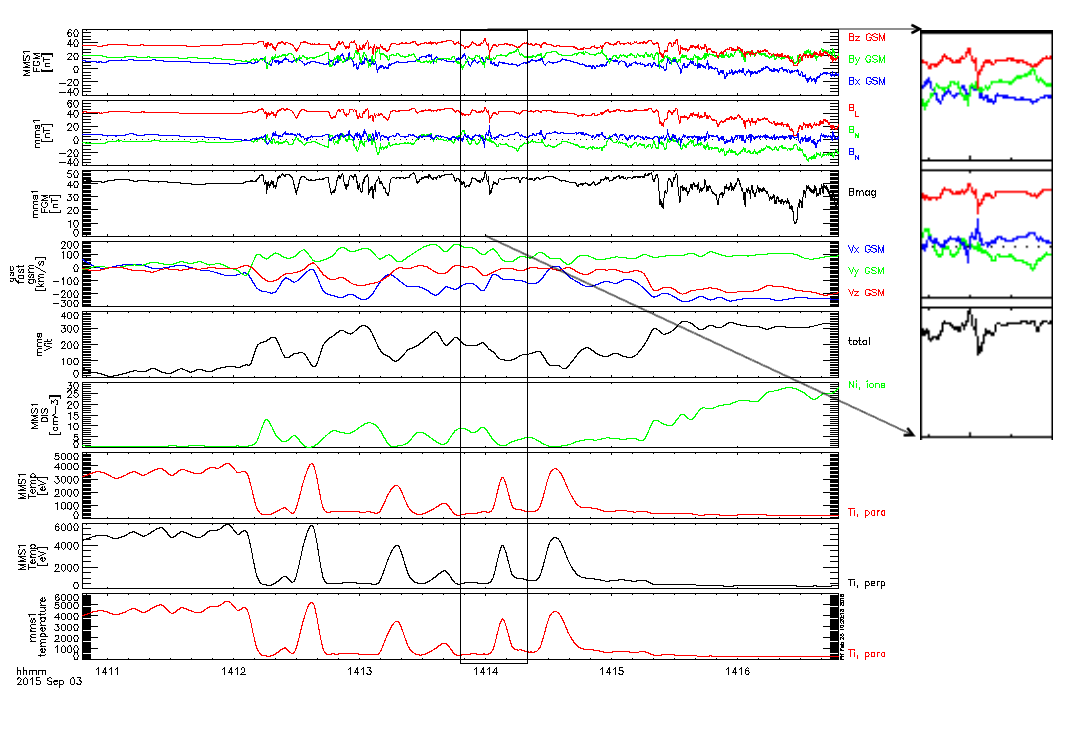} 
    \caption{An FTE Event. The image shows 15 variables in the following order $B_x$,$B_y$,$B_z$,$B_L$,$B_M$,$B_N$,$B_{mag}$,$V_x$,$V_y$,$V_z$,$V_{mag}$,$n_p$,$T_{par}$,$T_{per}$,$T_p$. The figure shows an example of flux transfer event (FTE), most visible due to its bipolar signature in $B_N$. $B_N$ is represented in second block in blue as shown in zoomed part. }
    \label{fig:fte}%
\end{figure}

Space physicists identify the FTEs in the data by first transforming the raw magnetic field data into the boundary normal coordinates based on the model of the Earth's magnetopause.  The three components of the magnetic field ($B_x,B_y,B_z$) are transformed into ($B_N, B_M, B_L$) where $B_N$ is the component along the magnetopause normal, $B_M$ is tangential to the magnetopause, and $B_L$ forms the third orthogonal coordinate. In this transformed frame, FTEs exhibit a bipolar signature in $B_N$ which makes it easier to identify the FTEs visually (Fig. \ref{fig:fte}).  It is important to note that it would be difficult to visually identify FTEs in the original frame as evident in Fig. \ref{fig:fte}. As such, this data set is ideal for testing and validation of our approach for identification of important channels. The most important channel for identification of FTEs is $B_N$ and the model should highlight that as such. For the relative importance of various variables to the classification of FTEs, we refer the reader to \citep{karimabadi2009new}

\sloppy Our data consists of 15 variables in the following order $B_x$,$B_y$,$B_z$,$B_L$,$B_M$,$B_N$,$B_{mag}$,$V_x$,$V_y$,$V_z$,$V_{mag}$,$n_p$,$T_{par}$,$T_{per}$,$T_p$. The $V$'s refer to components of the ion velocity in the original frame and its magnitude, $n$ is the plasma density, $T_{par},T_{per}$ refer to ion temperature parallel and perpendicular to the magnetic field, respectively, and $T_p$ refers to the total ion temperature. Data is labeled by whether a given time window has FTE events (class 1) or no events (class 0). We do not specify the beginning or end of the event. A given interval with FTEs may have one or more FTEs. The labels were created by space physicists through visual inspection of the data.

Data consists of 184 samples of class 0 and 227 samples of class 1 time series with an equal length of 5440.  This data is divided into 295 train samples (169 class 1's), validation size of 33 (20 class 1's), and test size of 83 (38 class 1's). During training we down sample the time-series from 5440 to 1440 for computational efficiency.

\begin{figure}[!t]    
\includegraphics[width=0.25\textwidth]{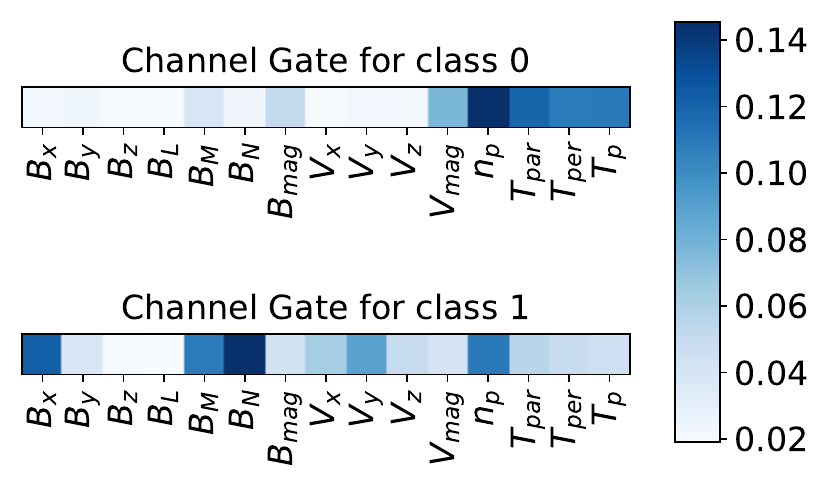} 
\caption{We perform average activation of the channel gate over the test set and then normalize the values by the test set size. As we can see, IETNet can separate variables by strongly picking first few channels which correspond to 2 body as shown in bottom panel. Likewise for class 0, the network is using all of the channels to classify it as a background class.}
\label{fig:mms-b}
 \end{figure}

In figure \ref{fig:mms-b} we show the channel localization by the model aggregated over the entire test set. The top and bottom bars show the aggregated channel localization for class 0 (no event) and class 1 (event), respectively. For class 1, the network has picked the magnetic field channels with the strongest importance given to $B_N$ as expected. Note that the second highest importance is given to $B_x$ which is the closest to $B_N$ in the original frame.  In class 0 cases, there would be nothing unique about $B_N$ or other magnetic field components and correctly the network has selected channels with plasma variables such as density and temperature as the most important.

\begin{figure}[!t]
\includegraphics[width=0.25\textwidth]{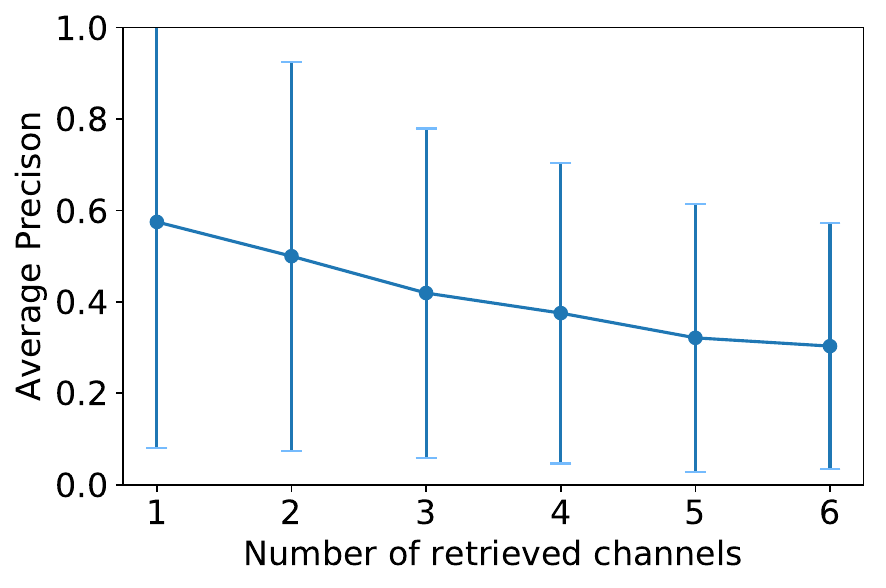} 
    
\caption{We plot mean average precision at various $k$'s. The model's agreement to the ground truth, specially at first few $k$'s, is evident. We have the six magnetic field channels as part of ground truth}%

\label{fig:mms-c}%
\end{figure}

\sloppy The importance of the magnetic field variables in class 1 events, as identified by the model, is further illustrated in \ref{fig:mms-c}.  We included $B_x,B_y,B_z,B_L,B_M,B_N,B_{mag}$ as a part of ground truth channels in case of class 1. As we can see the model has high precision when it retrieves top few channels and the score gradually decreases as we look at more number of channels. The trend shows high agreement of retrieved channels to ground truth channels. We also map the standard deviation of hit rate across test set of the retrieved channels to have a better understanding of model performance.

\section{Discussion}
\subsection{Classification Performance}
\begin{figure}[!t]
    \centering
    \includegraphics[width=0.3\textwidth]{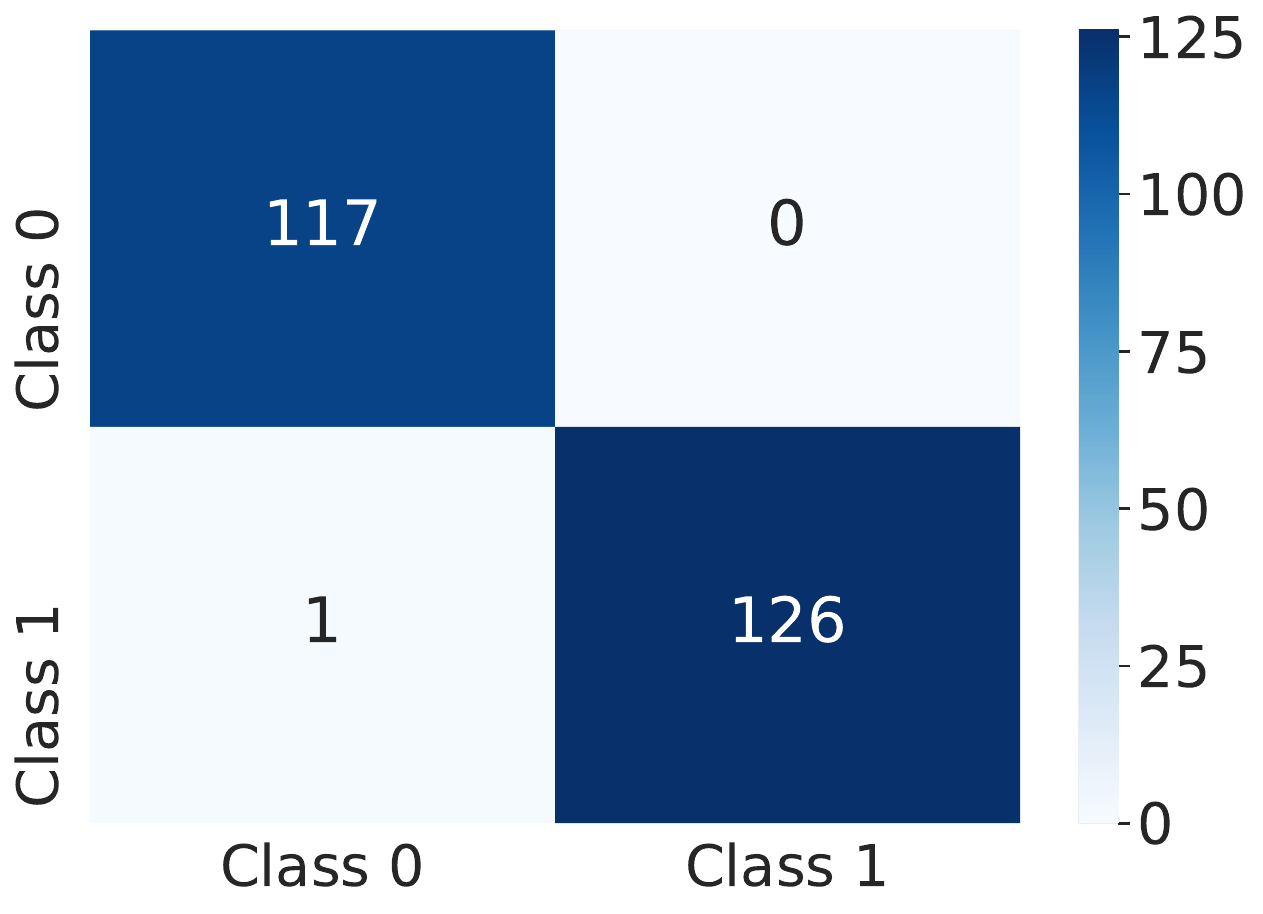} 
    \caption{Classification performance shows very high degree of agreement between ground truth and predicted labels for N-body simulation dataset.}
    
\label{fig:nbodya}
\end{figure}

For classification, we use ROC curves and confusion matrix. Classification performance of the IETNet on the N-body problem is shown using the confusion matrix in figure \ref{fig:nbodya}. The model is seen to have very high accuracy with only one misclassified example.

\par
\begin{figure}[!h]
    \centering
    \includegraphics[width=0.25\textwidth]{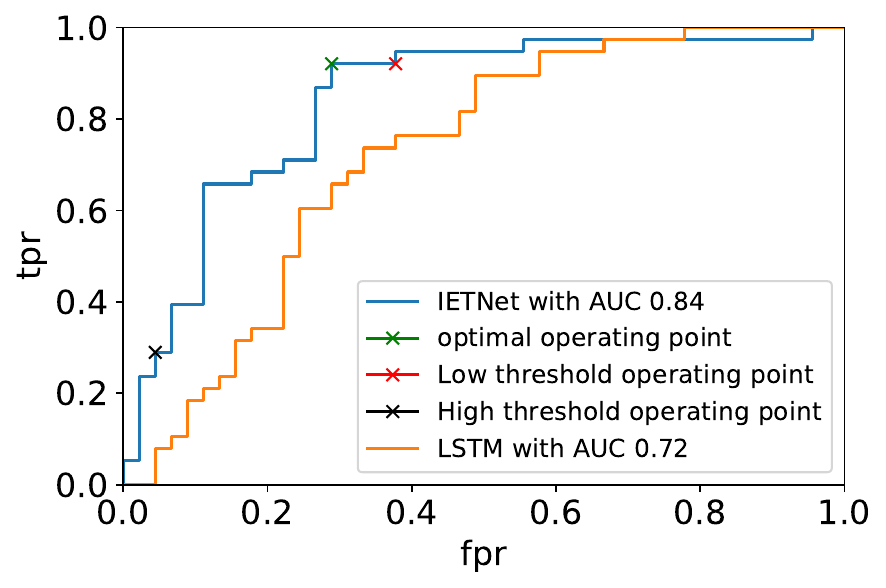}
    \caption{classification performance shows very high degree of agreement between ground truth and predicted labels for NASA data.}
\label{fig:mms-a}
\end{figure}

 In the first experiment on this data, we keep all 15 variables and then check whether the network selects $B_N$ as the most important channel. One can imagine that the accuracy of the classifier could impact the accuracy of the channel importance component. To disentangle this effect, we first plot the ROC of the classifier  on the test set. This is shown in figure \ref{fig:mms-a} where the optimal operating point is marked in green(obtained using validation set). The AUC is 0.84 and is significantly better than AUC of 0.72 for a standard LSTM.



\section{Conclusion}

Here we proposed a new neural network, IETNet, capable of identifying the most importance channels for each classification instance of multivariate time series data. The efficacy of this network was demonstrated through two examples, N body problem and in situ spacecraft measurements from a recent NASA mission.  Detailed  analysis of the model on N body simulation and NASA spacecraft sensor data reveals high degree of agreement between our prior knowledge of important channels and channels picked by the model. As most natural stimuli are time-continuous and multivariate, the approach promises to be of great utility in real-world applications. 
\section{Acknowledgements}
We thank Dimitry Fischer, Hudson Cooper, Jason Wilkes and Jonathan Driscoll for many useful discussions and comments during the course of this work. We also extend special thanks to Marcos Silveira and David Sibeck from NASA for sharing the data and providing insights regarding FTE events.